\documentclass[10pt,reqno]{article}
\usepackage{graphicx}
\usepackage{indentfirst,csquotes}

\usepackage{amssymb,amsthm,amsmath}
\usepackage{xcolor,paralist,hyperref,fancyhdr,etoolbox}

\usepackage{algpseudocode}
\usepackage{algorithm}

\usepackage{array}
\usepackage{multirow}
\usepackage{dcolumn}
\newcolumntype{2}{D{.}{}{2.0}}

\hypersetup{ colorlinks=true, linkcolor=black, filecolor=black, urlcolor=black }

\usepackage{lipsum}
\begin{document}
\title{Software Implementation of Digital Filtering via Tustin's Bilinear Transform} 
\author{Connor W. Herron}
\date{\today}
\maketitle

\begin{abstract}
The purpose of this work is to provide some notes on a software implementation for digital filtering via Tustin's Bilinear Transform. The first section discusses how to solve for the input/output coefficients by hand using a generalized approach called Horner's method. The second section presents some results of this generalized digital filtering approach using the IHMC Open Robotics Software (ORS) stack and Simulation Construction Set 2 (SCS2) \cite{ihmcORS, ihmcSCS}. This generalized approach can solve for the digital coefficients for any causal transfer function.
\end{abstract} 

\bigskip

\section{Introduction}
The purpose of this package is to build digital filters which can run in control loops at a consistent known frequency, $f_l$. We can represent filters in their continuous transfer function form:
\begin{equation}
    H(s) = \frac{Y(s)}{X(s)}= \frac{b_0s^{\rm m} + b_1s^{\rm m-1}+...+b_{\rm m-1}s + b_{\rm m}}{a_0s^{\rm n}+a_1s^{\rm n-1}+...+a_{\rm n-1}s + a_{\rm n}}
\end{equation}
where $\rm m \leq n$ for the transfer function to be causal. We are looking to convert the transfer function, $H(s)$, to a digital filter of the following form 
\begin{equation} \label{general_df}
    \begin{split}
        y_0 &= \rm \hat{b}_0y_1 + \hat{b}_1 y_{2}+...+\hat{b}_{\rm n-1}y_{\rm n-1} + \hat{a}_0x_0 + \hat{a}_1x_{1}+...+\hat{a}_{\rm n-1}x_{\rm n-1} + \hat{a}_{\rm n}x_{\rm n} \\
         &= \hat{\textbf{b}}^T\textbf{y} + \hat{\textbf{a}}^T \textbf{x}
    \end{split}
\end{equation}
where $\hat{\textbf{b}}, \textbf{y} \in \mathbb{R}^{\rm n-1 \times 1}$, $\hat{\textbf{b}} = \rm{col}[\hat{b}_0, \hat{b}_1, ... , \hat{b}_{\rm n-1}]$ and $\hat{\textbf{a}}, \textbf{x} \in \mathbb{R}^{\rm n \times 1}$, $\hat{\textbf{a}} = \rm{col}[\hat{a}_0, \hat{a}_1, ... , \hat{a}_{\rm n-1}, \hat{a}_{\rm n}]$ where $\hat{b}_i, \hat{a}_i \in \mathbb{R}$ are scalars with $0 \leq i \leq n$ refers to the output order. The vectors $\textbf{x} = \rm{col}[x_0, x_{\rm 1}, ..., x_{\rm n-1}, x_{\rm n}]$ and $\textbf{y} = \rm{col}[y_1, y_{\rm 2}, ..., y_{\rm n-1}]$ represent the input/output ``history" where the subscript represents the loop iteration (eg. $x_0$ is the current input, $x_1$ is the previous input, $x_2$ is the previous previous input, etc.).
Equation (\ref{general_df}) must be solved in a loop running at a consistent frequency, $f_l$, where the input/output histories must be updated. Failure to achieve a consistent frequency will no longer represent the desired transfer function. The input/output coefficient vectors $\hat{\textbf{a}},\, \hat{\textbf{b}}$ are constant throughout the control sequence and only need to be solved during initialization. \textbf{The goal of this work is to describe a general approach which solves for the digital representation of any continuous, causal transfer function, which translates to solving for a set of input/output coefficients, $\hat{\textbf{a}},\, \hat{\textbf{b}}$ to be used in Eq. (\ref{general_df}) as part of a control loop.}

\section{Examples}
The digital transformation used here is Tustin's Bilinear Transform method, which simply represents the continuous frequency variable 
$$ s = f_l \cdot ln(z) = 2f_c \cdot \big[\frac{z-1}{z+1} + \frac{1}{3}\cdot (\frac{z-1}{z+1})^3 + \frac{1}{5}\cdot (\frac{z-1}{z+1})^5 + \frac{1}{7}\cdot (\frac{z-1}{z+1})^7 \big] $$
\begin{equation} \label{tustinsTransform}
    s = 2f_l \frac{z-1}{z+1}
\end{equation}
where $f_l$ is the digital loop frequency at which Eq. (\ref{general_df}) is solved at. After solving the example using the general method, I'll show an alternative approach which is more software friendly.
\subsection{General Method} 

\subsubsection{Example 1: General Method} 

This first example is a simple 1st order, unity-gain, low-pass transfer function of the form
$$ H(s) = \frac{1}{\frac{s}{\omega_0} + 1}, \quad \omega_0 = 0.1 \,\text{rad/s} $$
\begin{equation} \label{eq.1_gm}
    H(s) = \frac{1}{10s + 1}
\end{equation}
where we want to solve for a set of digital input/output coefficient vectors $\hat{\textbf{a}}, \hat{\textbf{b}}$ from Eq. (\ref{general_df}). \\\\
If we have a loop frequency of $f_l = 0.1$ Hz (chosen simply chosen for mathematical convenience, and is considerably lower than normal loop speeds), we can substitute Eq. (\ref{tustinsTransform}) in (\ref{eq.1_gm}) to get
\begin{align*}
    H[z] &= \frac{1}{10\cdot(2\cdot0.1 \frac{z-1}{z+1}) + 1} \\
    H[z] &= \frac{z+1}{2\cdot(z-1)+z+1} \\
    H[z] &= \frac{Y[z]}{X[z]} = \frac{z+1}{3z-1} \\
\end{align*}
where we can split the transfer function into $Y[z]$ and $X[z]$ components
\begin{align*}
    Y[z](3z-1) &= X[z](z+1) \\
    Y[z](1-\frac{1}{3}z^{-1}) &= X[z] (\frac{1}{3} + \frac{1}{3}z^{-1})
\end{align*}
Finally, we can derive our digital filter 
\begin{equation}
    y_0 = \frac{1}{3}y_{1} + \frac{1}{3}x_{0} + \frac{1}{3}x_{1}
\end{equation}
where our set of input/output coefficient vectors is
\begin{equation} \label{generalSol1}
\boxed{
    \hat{\textbf{a}}^T = [\frac{1}{3},\, \frac{1}{3}], \quad \text{and} \quad \hat{\textbf{b}}^T = [\frac{1}{3}]}
\end{equation}

While this approach is relatively simple to complete by hand, it's quite difficult to generalize to more difficult examples. In this next example, the desired transfer function is 2nd order.

\subsubsection{Example 2: General Method} 
The 2nd order low-pass, Butterworth is a common filter of the form
\begin{equation}
    H(s) = \frac{\omega_c^2}{s^2+\sqrt{2}\omega_cs + \omega_c^2}
\end{equation}

For simple numerical values, I'll choose $\omega_c = \sqrt{2}$ rad/s, with $f_l = 1$ Hz. We can start by immediately plugging in Eq. (\ref{tustinsTransform}) where
\begin{align*}
    H(s) &= \frac{2}{s^2+2s+2} \\
    H[z] &= \frac{2}{(2f_l\frac{z-1}{z+1})^2 + 2(2f_l\frac{z-1}{z+1}) + 2} \\
    H[z] &= \frac{2(z+1)^2}{(4f_l^2(z-1)^2 + 4f_l(z-1)(z+1)) +2(z+1)^2} \\
    H[z] &= \frac{2(z^2+2z+1))}{(4(z^2-2z+1) + 4(z^2-1)+2(z^2+2z+1)} \\
    H[z] &= \frac{2z^2+4z+2}{4z^2-8z+4 + 4z^2-4+2z^2+4z+2} \\
    H[z] &= \frac{Y[z]}{X[z]} = \frac{2z^2+4z+2}{10z^2-4z+2} 
\end{align*}
where we can finally split the transfer function into $Y[z]$ and $X[z]$ components
\begin{align*}
    Y[z](10z^2-4z+2) &= X[z](2z^2+4z+2) \\
    Y[z](1-0.4z^{-1} + 0.2z^{-2}) &= X[z](0.2+0.4z^{-1} + 0.2z^{-2})
\end{align*}
Finally, our digital filter can be derived as 
\begin{equation}
    y_0 = 0.4y_{1} -0.2y_{2} + 0.2x_{0} + 0.4x_{1} +0.2x_{2}
\end{equation}
and the set of input/output coefficient vectors is
\begin{equation} \label{io_coeff_1}
\boxed{
    \hat{\textbf{a}} = [0.2,\, 0.4,\, 0.2],\quad \text{and} \quad 
    \hat{\textbf{b}} = [0.4,\, -0.2]}
\end{equation}

Using the general method, it can be hard to write an algorithm to find $\hat{\textbf{a}},\, \hat{\textbf{b}}$ for any transfer function. In addition, a generalized method could be computational expensive, which isn't a deal breaker since this task would be completed during initialization prior to loop running, but we may have hundreds of filters and it would be good to consider efficiency. 

\subsection{Horner's Method} 

Horner's method is a convienent way of solving the $H(s) \rightarrow H[z]$ conversion in a stepwise manner which is mathematically equivalent and software generalizable to any causal transfer function \cite{davies1974bilinear}. Horner's method relies on the fact that we can rewrite Tustin's bilinear transform as
\begin{equation}
    \begin{split}
        \frac{1}{s} &= \frac{1}{2f_l}\cdot \frac{z+1}{z-1} \\
        &= \frac{1}{2f_l}\cdot \frac{z+1-1+1}{z-1} \\
        &= \frac{1}{2f_l}\cdot (\frac{2}{z-1} + 1)
    \end{split}
\end{equation}

In this method, instead of simply substituting Eq. (\ref{tustinsTransform}) into the transfer function, this method retains the polynomial orders of power while separately considering the numerator and denominator. The method will convert 
\begin{align}
    \begin{split}
        F(s) \rightarrow F(x) \rightarrow F(x+1) \rightarrow F(\frac{1}{x} + 1) &\rightarrow F(\frac{2}{x} + 1) 
        \rightarrow F(\frac{2}{x-1} +1) \\
        & \equiv F\left(\frac{1}{2f_l}\cdot(\frac{2}{z-1}+1)\right)=F[z]
    \end{split}
\end{align}

\textbf{Note:} The variable $x$ is a general variable to transition between layers of the $F(s) \rightarrow F[z]$, where $x$ is not equivalent between layers. In addition, the multiplication of $2f_l$ can take place in an step in transition where I've chosen to the first step where $F(s) \rightarrow F(x)$. In addition, there are steps here where the same multiplication will take place across the numerator and denominator for convenience while being mathematically equivalent.

\subsubsection{Example 1: Horner's Method}

This first example is a simple 1st order, unity-gain, low-pass transfer function of the form
$$ H(s) = \frac{1}{\frac{s}{\omega_0} + 1}, \quad \omega_0 = 0.1 \,\text{rad/s} $$
\begin{equation} \label{eq.1_hm}
    H(s) = \frac{1}{10s + 1}
\end{equation}
where $f_l = 0.1$ Hz and we want to solve for a set of digital input/output coefficient vectors $\hat{\textbf{a}}, \hat{\textbf{b}}$ from Eq. (\ref{general_df}). \\

1) $\boldsymbol{[F(s) \rightarrow F(x)]}$ We can start by separating the numerator and denominator, divide the polynomial by $s^n$, then substitute $s = \frac{2 f_l}{x}$. As previously mentioned, you can choose to multiply by $2 f_l$ at any step in this transition.
\begin{align*}
    \underline{\textbf{Nume}}&\underline{\textbf{rator}} & \underline{\textbf{Deno}}&\underline{\textbf{minator}}  \\
    N(s) &= 1 & D(s) &= 10s+1 \\
    \frac{N(s)}{s} &= \frac{1}{s} & \frac{D(s)}{s} &= 10 + \frac{1}{s} \\
    \frac{N(s)}{s} \rightarrow N(x) &= \frac{x}{2f_l} & \frac{D(s)}{s} \rightarrow D(x) &= 10 + \frac{x}{2 f_l} \\
    N(x) &= \frac{x}{0.2} & D(x) &= 10 + \frac{x}{0.2} \\
    N(x) &= 5x & D(x) &= 10 + 5x 
\end{align*}

2) $\boldsymbol{F(x) \rightarrow F(x+1)]}$ Next we can decrease the zeros by 1 using synthetic division where \\
\begin{align*}
    \underline{\textbf{Nume}}&\underline{\textbf{rator}} & \underline{\textbf{Deno}}&\underline{\textbf{minator}}  \\
    N(x) &= 5x & D(x) &= 5x + 10  \\
    &
    \begin{array}{rrr}
        & 5 & 0 \\
        + & \downarrow & 5 \\\cline{2-3}
         & \downarrow & {|} 5 \\\cline{2-3}
        & 5 &   
    \end{array} & & 
    \begin{array}{rrr}
        & 5 & 10 \\
        + & \downarrow & 5 \\\cline{2-3}
         & \downarrow & {|} 15 \\\cline{2-3}
        & 5 &   
    \end{array} \\
    N(x) \rightarrow N(x+1) &= 5x + 5 & D(x) \rightarrow D(x+1) &= 5x + 15 
\end{align*}
\newpage
3) $\boldsymbol{[F(x+1) \rightarrow F(\frac{1}{x} + 1)]}$ Replace all zeros w/ reciprocals, equivalent to flipping the coefficients.

\begin{align*}
    \underline{\textbf{Nume}}&\underline{\textbf{rator}} & \underline{\textbf{Deno}}&\underline{\textbf{minator}}  \\
    N(x+1) &= 5x + 5 & D(x+1) &= 5x + 15 \\
    N(\frac{1}{x}+1) &= 5x+5 & D(\frac{1}{x} + 1) &= 15x + 5
\end{align*}

4) $\boldsymbol{F(\frac{1}{x}+1) \rightarrow F(\frac{2}{x}+1)]}$ Scale the polynomial zeroes by 2. Note can you could scale by either $\frac{1}{2}$ or $2$ for the orders of power since the polynomials are in both the numerator and denominator.

\begin{align*}
    \underline{\textbf{Nume}}&\underline{\textbf{rator}} & \underline{\textbf{Deno}}&\underline{\textbf{minator}}  \\
    N(\frac{1}{x}+1) &= 5x+5 & D(\frac{1}{x}+1) &= 15x + 5 \\
    N(\frac{2}{x}+1) &= 5(\frac{x}{2}) + 5 & D(\frac{2}{x} + 1) &= 15(\frac{x}{2}) + 5 \\
    &= 2.5x + 5 & &= 7.5x + 5
\end{align*}

5) $\boldsymbol{F(\frac{2}{x}+1) \rightarrow F(\frac{2}{x-1}+1)]}$ Increase all polynomial zeros by 1 using synthetic division.
\begin{align*}
    \underline{\textbf{Nume}}&\underline{\textbf{rator}} & \underline{\textbf{Deno}}&\underline{\textbf{minator}}  \\
    N(\frac{2}{x}+1) &= 2.5x + 5 & D(\frac{2}{x}+1) &= 7.5x + 5  \\
    &
    \begin{array}{rrr}
        & 2.5 & 5 \\
        - & \downarrow & 2.5 \\\cline{2-3}
         & \downarrow & {|} 2.5 \\\cline{2-3}
        & 2.5 &   
    \end{array} & & 
    \begin{array}{rrr}
        & 7.5 & 5 \\
        - & \downarrow & 7.5 \\\cline{2-3}
         & \downarrow & {|} -2.5 \\\cline{2-3}
        & 7.5 &   
    \end{array} \\
   N(\frac{2}{x}+1) \rightarrow N(\frac{2}{x-1}+1) &= 2.5x + 2.5 & D(\frac{2}{x}+1) \rightarrow D(\frac{2}{x-1}+1) &= 7.5x - 2.5 
\end{align*}

6) $\boldsymbol{F(x) = F[z]]}$ Substitute in $x=z$ to equivalently write the transfer function as
\begin{equation}
    H[z] = \frac{Y[z]}{X[z]} = \frac{N[z]}{D[z]} = \frac{2.5z + 2.5}{7.5z -2.5}
\end{equation}

7) Derive the digital filter where
\begin{align*}
    Y[z](7.5z-2.5) &= X[z](2.5z+2.5) \\
    Y[z](1-\frac{1}{3}z^{-1}) &= X[z](\frac{1}{3} + \frac{1}{3}z^{-1})
\end{align*}
where finally we have our digital filter.
\begin{equation}
    y_0 = \frac{1}{3}y_{1} + \frac{1}{3}x_0 + \frac{1}{3}x_{1}
\end{equation}
with input/output coefficients of
\begin{equation}
\boxed{
    \hat{\textbf{a}}^T = [\frac{1}{3},\, \frac{1}{3}], \quad \text{and} \quad \hat{\textbf{b}}^T = [\frac{1}{3}]}
\end{equation}
Note that this solution matches the solution from the general approach shown in (\ref{generalSol1}).

\subsubsection{Example 2: Horner's Method}
The 2nd order low-pass, Butterworth is a common filter of the form
\begin{equation}
    H(s) = \frac{\omega_c^2}{s^2+\sqrt{2}\omega_cs + \omega_c^2}
\end{equation}

For simple numerical values, I'll choose $\omega_c = \sqrt{2}$ rad/s, with $f_l = 1$ Hz. So our transfer function is of the form
\begin{equation*}
    H(s) = \frac{2}{s^2 + 2s+2}
\end{equation*}

1) $\boldsymbol{[F(s) \rightarrow F(x)]}$ Separate the numerator and denominator, divide each polynomial by $s^n$, and then substitute $s = \frac{2f_l}{x}$.

\begin{align*}
    \underline{\textbf{Nume}}&\underline{\textbf{rator}} & \underline{\textbf{Deno}}&\underline{\textbf{minator}}  \\
    N(s) &= 2 & D(s) &= s^2 + 2s+2 \\
    \frac{N(s)}{s^2} &= \frac{2}{s^2} & \frac{D(s)}{s^2} &= 1 + \frac{2}{s} + \frac{2}{s^2} \\
   \frac{N(s)}{s^2} \rightarrow N(x) &= \frac{x^2}{2} & \frac{D(s)}{s^2} \rightarrow D(x) &= 1 + x + \frac{x^2}{2} \\
    N(s) \rightarrow N(x) &= 0.5x^2 & D(s) \rightarrow D(x) &= 0.5x^2 + x+ 1 \\
\end{align*}

2) $\boldsymbol{[F(x) = F(x+1)]}$ Decrease the zeros by 1 using synthetic division
\begin{align*}
    \underline{\textbf{Nume}}&\underline{\textbf{rator}} & \underline{\textbf{Deno}}&\underline{\textbf{minator}}  \\
    N(x) &= 0.5x^2  & D(x) &= 0.5x^2+x+1  \\
    &
    \begin{array}{rrrr}
          & 0.5 & 0 & 0 \\
        + & \downarrow & 0.5 & 0.5 \\\cline{2-4}
          & 0.5 & 0.5 & {|} 0.5 \\\cline{4-4}
        + & \downarrow & 0.5 \\\cline{2-3}
        + & 0.5 & {|} 1 \\\cline{2-3}
          & 0.5 &   
    \end{array} & & 
    \begin{array}{rrrr}
          & 0.5 & 1 & 1 \\
        + & \downarrow & 0.5 & 1.5 \\\cline{2-4}
          & 0.5 & 1.5 & {|} 2.5 \\\cline{4-4}
        + & \downarrow & 0.5 \\\cline{2-3}
        + & 0.5 & {|} 2 \\\cline{2-3}
          & 0.5 & 
    \end{array} \\
    N(x+1) &= 0.5x^2+x+0.5 & D(x+1) &= 0.5x^2+2x+2.5 
\end{align*}

3) $\boldsymbol{[F(x+1) = F(\frac{1}{x}+1)]}$ Replace all zeros w/ reciprocals, which is equivalent to flipping the coefficients

\begin{align*}
    \underline{\textbf{Nume}}&\underline{\textbf{rator}} & \underline{\textbf{Deno}}&\underline{\textbf{minator}}  \\
   N(x+1) &= 0.5x^2+x+0.5 & D(x+1) &= 0.5x^2+2x+2.5 \\
   N(\frac{1}{x} + 1) &= 0.5x^2 +x + 0.5 & D(\frac{1}{x}+1) &= 2.5x^2 + 2x + 0.5 
\end{align*}

4) $\boldsymbol{[F(\frac{1}{x}+1) = F(\frac{2}{x}+1)]}$ Scale the polynomial zeros by 2 (or $\frac{1}{2}$).

\begin{align*}
    \underline{\textbf{Nume}}&\underline{\textbf{rator}} & \underline{\textbf{Deno}}&\underline{\textbf{minator}}  \\
    N(\frac{1}{x} + 1) &= 0.5x^2+x+0.5 & D(\frac{1}{x} + 1) &= 2.5x^2 + 2x + 0.5 \\ 
   N(\frac{2}{x} + 1) &= 0.5(\frac{x}{2})^2 +(\frac{x}{2}) + 0.5 & D(\frac{2}{x}+1) &= 2.5(\frac{x}{2})^2 + 2(\frac{x}{2}) + 0.5 \\
   N(\frac{2}{x} + 1) &= 0.125x^2 + 0.5x + 0.5 & D(\frac{2}{x} + 1) &= 0.625x^2 +x+0.5
\end{align*}

5) $\boldsymbol{[F(\frac{2}{x}+1) = F(\frac{2}{x-1}+1)]}$ Increase all polynomial zeros by 1 using synthetic division.
\begin{align*}
    \underline{\textbf{Nume}}&\underline{\textbf{rator}} & \underline{\textbf{Deno}}&\underline{\textbf{minator}}  \\
    N(x) &= 0.125x^2 + 0.5x + 0.5  & D(x) &= 0.625x^2 +x+0.5  \\
    &
    \begin{array}{rrrr}
          & 0.125 & 0.5 & 0.5 \\
        - & \downarrow & 0.125 & 0.375 \\\cline{2-4}
          & 0.125 & 0.375 & {|} 0.125 \\\cline{4-4}
        - & \downarrow & 0.125 \\\cline{2-3}
        - & 0.125 & {|} 0.25 \\\cline{2-3}
          & 0.125 &   
    \end{array} & & 
    \begin{array}{rrrr}
          & 0.625 & 1 & 0.5 \\
        - & \downarrow & 0.625 & 0.375 \\\cline{2-4}
          & 0.625 & 0.375 & {|} 0.125 \\\cline{4-4}
        - & \downarrow & 0.625 \\\cline{2-3}
        - & 0.625 & {|} -0.25 \\\cline{2-3}
          & 0.625 & 
    \end{array} \\
    N(x) &= 0.125x^2+0.25x+0.125 & D(x) &= 0.625x^2-0.25x+0.125 
\end{align*}

6) $\boldsymbol{[F(x) = F[z]]}$ Substitute in $x=z$ and equivalently write the digital transfer function as
\begin{equation*}
    H[z] = \frac{Y[z]}{X[z]} = \frac{N[z]}{D[z]} = \frac{0.125z^2 + 0.25z +0.125}{0.625z^2 -0.25z +0.125}
\end{equation*}

7) Derive the digital filter.
\begin{align*}
    Y[z](0.625z^2 -0.25z +0.125) &= X[z](0.125z^2 + 0.25z +0.125) \\
    Y[z](1-0.4z^{-1}+0.2z^{-2}) &= X[z](0.2+0.4z^{-1}+0.2z^{-2})
\end{align*}
Finally, our digital filter is
\begin{equation}
    y_0 = 0.4y_{1}-0.2y_{2}+0.2x_0+0.4x_{1}+0.2x_{2}
\end{equation}
where the digital input/output coefficients are
\begin{equation}
\boxed{
   \hat{\textbf{a}}^T = [0.2,\, 0.4,\, 0.2], \quad \text{and} \quad \hat{\textbf{b}}^T = [0.4,\, -0.2]}
\end{equation}

Notes that this solution is equivalent to that in the general method in (\ref{io_coeff_1}).

\newpage

\section{Results}
  \subsection{Chirp Signals}
I'll show some comparisons of transfer functions using the intended model and results of an input/output response of that model's filter with an exponential chirp input. A chirp signal is a sinusoid whose frequency changes  \textit{linearly} or \textit{exponentially} with time depending on the desired waveform. This type of signal can be used to sweep through a frequency spectrum for generating a bode plot of the linear system. Personally, I prefer using exponential chirp signals so that the number of data points spent on each frequency matches the log-scale graph of the bode plot (i.e. we get more data points on lower-frequencies that we usually care more about). Alternatively, one could use a sequence of sinusoids instead of a dynamic sinusoid. That allows you to separate the transient and steady state responses, which is otherwise impossible to separate in a chirp signal response.

\vspace{0.5cm}
Here's the equation of the frequency change, $\omega_{\rm c}(t)$, with respect to time for linear and exponential chirp signals
\\
\begin{equation} \label{lin_chirp}
    \omega_{\rm c}^{\rm lin}(t) = \left(\frac{\omega_{\rm max} - \omega_{\rm min}}{T}\right)\cdot t + \omega_{\rm min}
\end{equation}
\\
\begin{equation} \label{exp_chirp}
    \omega_{\rm c}^{\rm exp}(t) = \omega_{\rm min}\cdot \left(\frac{\omega_{\rm max}}{\omega_{\rm min}}\right)^{\frac{t}{T}}
\end{equation}

\begin{figure}[h] 
\includegraphics[width=\linewidth]{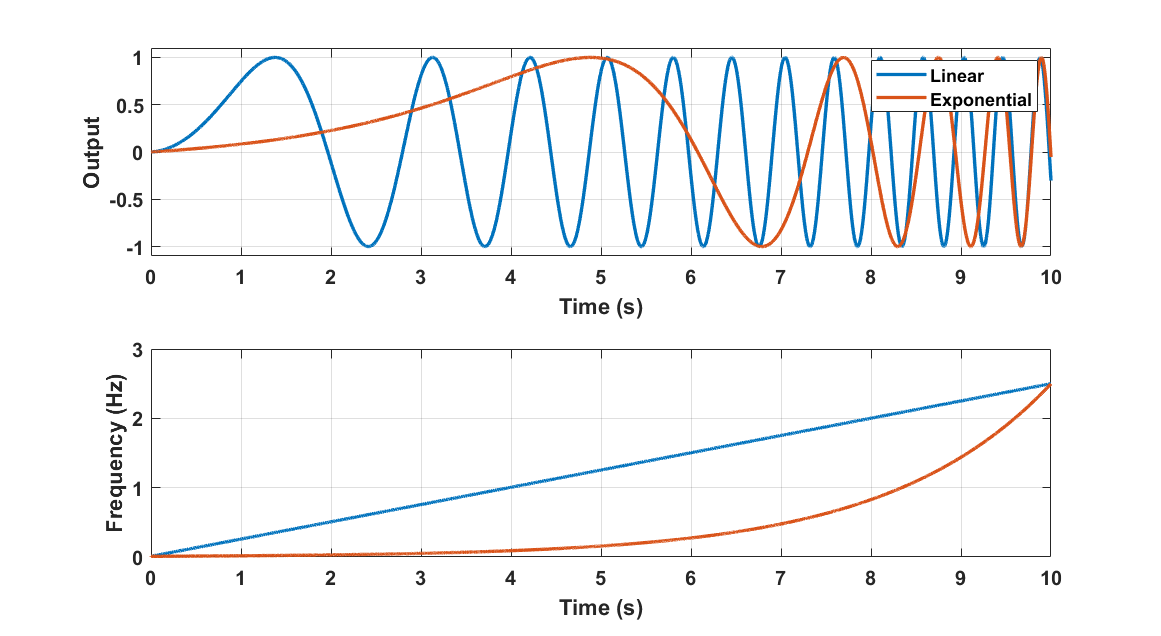}
\centering
\caption{Linear and Exponential Chirp Waveforms from 0.01 to 2.5 Hz with amplitude of 1}
\label{chirp_plot}
\end{figure}

Where $\omega_{\rm min}$ is the minimum frequency in rad/s, $\omega_{\rm max}$ is the maximum frequency in rad/s, and $T$ is total time in seconds. Using (\ref{lin_chirp}) and (\ref{exp_chirp}), the chirp signal for a digital system can be represented as
\begin{align}
    \cos_i &= \cos(\omega_c[t]\cdot \Delta t) \cos_{i-1} - \sin(\omega_c[t]\cdot \Delta t) \sin_{i-1} \\
    \sin_i &= \sin(\omega_c[t]\cdot \Delta t) \cos_{i-1} + \cos(\omega_c[t]\cdot \Delta t) \sin_{i-1}
\end{align}    
Finally, the chirp equation is represented by the following
\begin{equation}
    x_i = A\sin_i
\end{equation}
where $A$ refers to the amplitude. Figure \ref{chirp_plot} shows a plot of linear and exponential chirp signals where $\omega_{\rm min} = 2\pi\cdot0.01$ rad/s, $\omega_{\rm max} = 2\pi\cdot 2.5$ rad/s, $T = 10$ sec, and $A = 1$. 

\subsection{Software Implementation of Digital Filtering}
The algorithm for generating and using the filter in a control loop can be summarized by the following pseudocode:

\begin{algorithm} \label{algorithm}
\caption{Using a filter in a control loop.}\label{alg:cap}
\begin{algorithmic}
\item{\textbf{\% Solve for Digital Coefficients, }$\hat{\textbf{a}}$ \textbf{and} $\hat{\textbf{b}}$\textbf{, from Transfer Function,} $H_s(s) = \frac{N_s(s)}{D_s(s)}$.}
\State \text{Check }$n \geq m$
\State $\text{N}_z = \text{solve}_{\text{Fs}\rightarrow\text{Fz}}(\text{N}_s)$
\State $\text{D}_z = \text{solve}_{\text{Fs}\rightarrow\text{Fz}}(\text{D}_s)$
\State $\text{Scale } \text{N}_z \text{ and } \text{D}_z \text{ coefficients by } \text{D}_z\text{[0].}$
\State $\text{Multiply }\text{D}_z\text{ coefficients by -1 and solve for the } \hat{\textbf{a}} \text{ and } \hat{\textbf{b}} \text{ column vectors.}$ \\

\State \text{firstTick = true}\\

\item{\textbf{\% Run Control Loop.}}
\While{Control Loop is active}
\State $\textbf{\% Initialize output and input history vectors to first input.}$
\If{\text{firstTick}}
    \State $\textbf{y} = x_0\cdot \textbf{col}(1)_{\rm (n-1) \times 1}$ \Comment{Heuristic for avoiding high outputs during initialization.}
    \State $\textbf{x} = x_0\cdot \textbf{col}(1)_{\rm n \times 1}$
    \State $\text{firstTick = false}$
\EndIf 
\\

\item{\textbf{\% Update input vector, where $\textbf{x} \in \mathbb{R}^{\rm n \times 1}$.}}
\State $\textbf{x} = [x_0,\, x_1,\, ...\, x_{\rm n-1}, x_{\rm n}]^T$
\\

\item{\textbf{\% Solve for next filtered output.}}
\State $y_0 = \hat{\textbf{b}}^T\textbf{y} + \hat{\textbf{a}}^T \textbf{x}$

\\
\item{\textbf{\% Update output vector, where $\textbf{y} \in \mathbb{R}^{\rm n-1 \times 1}$.}}
\State $\textbf{y} = [y_0,\, y_1,\,...\, y_{\rm n-2},\, y_{\rm n-1}]^T$

\EndWhile
\end{algorithmic}
\end{algorithm}

The following pages show some examples of the digital filtering performance compared to an ideal model. This experimentation is completed using IHMC's Open Robotics Software (ORS) \cite{ihmcORS} and Simulation Construction Set 2 (SCS2) \cite{ihmcSCS}.

\newpage

\subsubsection{1st Order Low-Pass Filter:} 
The following transfer function:
\begin{equation*}
    H(s) = \frac{1}{\frac{1}{\tau}s+1}
\end{equation*}
where $\tau = 2\pi \cdot 10$. This transfer function can be digitally represented by the following input/output coefficient vectors with $f_l = 1000$ Hz.
\begin{align*}
    \hat{\textbf{a}} &= [3.0459E-02, 3.0459E-02] \\
    \hat{\textbf{b}} &= [9.3908E-01]
\end{align*}

The input/output and bode plot from this digital filter are as follows.
\begin{figure}[h] 
\includegraphics[width=0.9\linewidth]{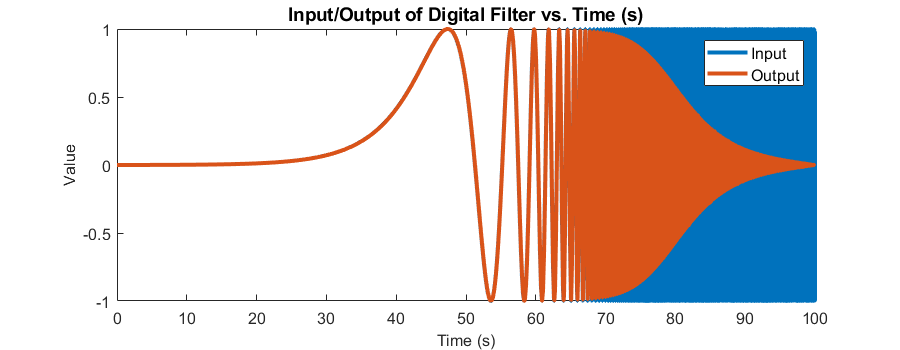}
\centering
\vspace{-0.25cm}
\caption{Input-Output plot of the 1st Order LP Filter with an exponential chirp input of 1 amplitude from 0.1 to 100 Hz}
\label{IO_Order1LP}
\end{figure}

\begin{figure}[h] 
\includegraphics[width=0.9\linewidth]{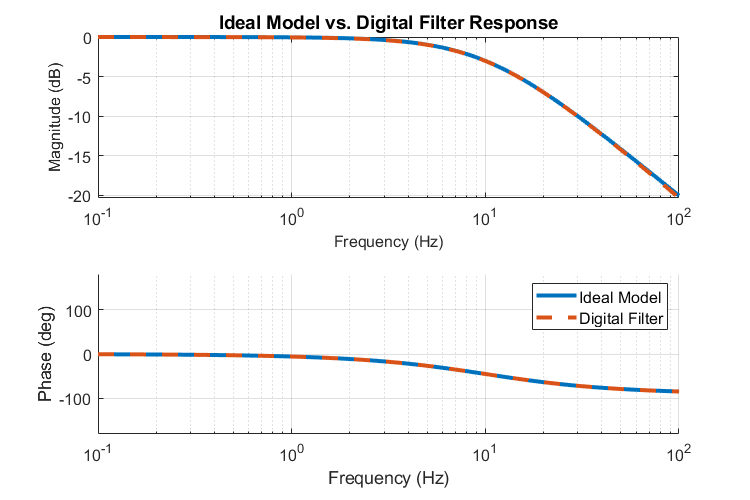}
\centering
\vspace{-0.25cm}
\caption{Bode Plot of filter with an exponential chirp input of 1 amplitude from 0.1 to 100 Hz}
\label{Bode_Order1LP}
\end{figure}

\subsubsection{2nd order Low-Pass Butterworth Filter:} \label{sec:order2lpbutter_filter}
The following transfer function:
\begin{equation*}
    H(s) = \frac{\omega_c^2}{s^2 + \sqrt{2}\omega_cs + \omega_c^2}
\end{equation*}
where $\omega_c = 2\pi \cdot 10$. This transfer function can be digitally represented by the following input/output coefficient vectors with $f_l = 1000$ Hz.
\begin{align*}
    \hat{\textbf{a}} &= [9.4408E-04,  1.8882E-03,  9.4408E-04] \\
    \hat{\textbf{b}} &= [1.9112E+00, -9.1500E-01]
\end{align*}

The input/output and bode plot from this digital filter are as follows.
\begin{figure}[h] 
\includegraphics[width=0.9\linewidth]{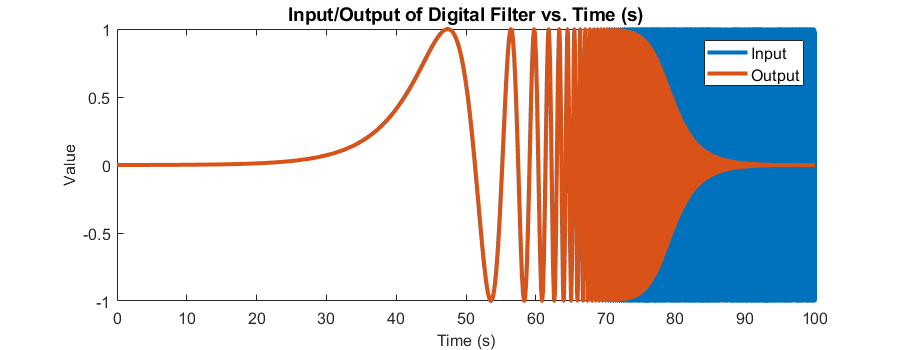}
\centering
\vspace{-0.25cm}
\caption{Input-Output plot of 2nd order LP Butterworth filter with an exponential chirp input of 1 amplitude from 0.1 to 100 Hz}
\label{IO_Order2LPButter}
\end{figure}

\begin{figure}[h] 
\includegraphics[width=0.9\linewidth]{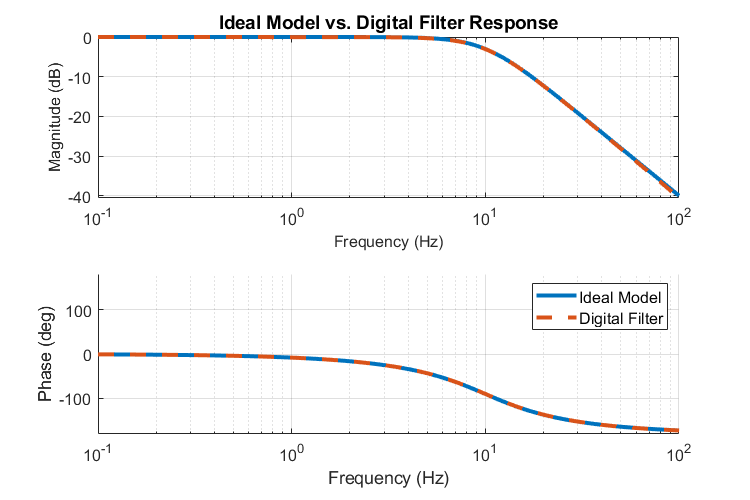}
\centering
\vspace{-0.25cm}
\caption{Bode Plot of filter with an exponential chirp input of 1 amplitude from 0.1 to 100 Hz}
\label{Bode_Order2LPButter}
\vspace{-1cm}
\end{figure}

\subsubsection{Notch Filter:}
The following transfer function:
\begin{equation*}
    H(s) = \frac{s^2 + \omega_n^2}{s^2 + \frac{\omega_n}{Q}s + \omega_n^2}
\end{equation*}
where $\omega_n = 2\pi \cdot 60$ and the quality factor, $Q = 5$. This transfer function can be digitally represented by the following input/output coefficient vectors with $f_l = 1000$ Hz.
\begin{align*}
    \hat{\textbf{a}} &= [9.6487E-01, -1.7973E+00,  9.6487E-01] \\
    \hat{\textbf{b}} &= [1.7973E+00, -9.2975E-01]
\end{align*}

The input/output and bode plot from this digital filter are as follows.
\begin{figure}[h] 
\includegraphics[width=0.9\linewidth]{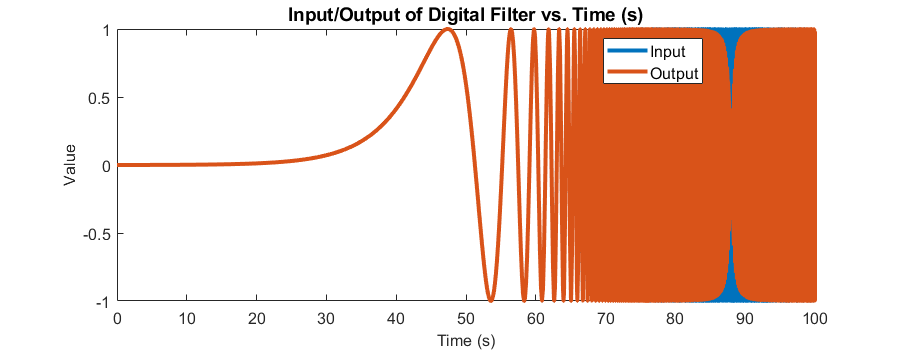}
\centering
\vspace{-0.25cm}
\caption{Input-Output plot of Notch Filter set to 60 Hz with an exponential chirp input of 1 amplitude from 0.1 to 100 Hz}
\label{IO_Notch}
\end{figure}

\begin{figure}[h] 
\includegraphics[width=0.9\linewidth]{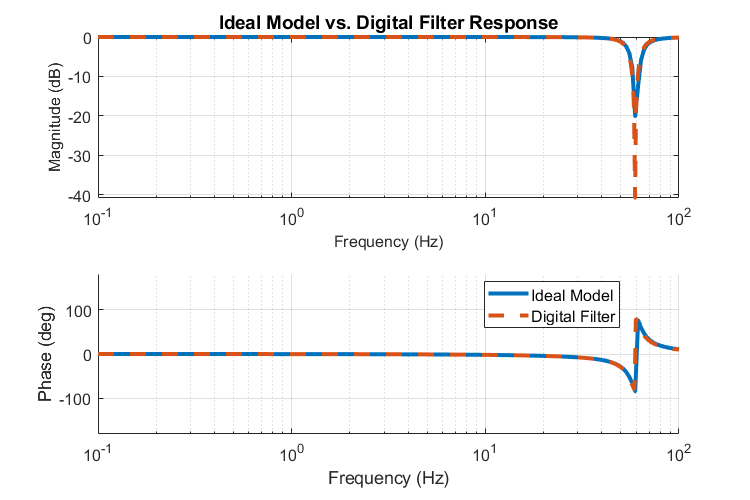}
\centering
\vspace{-0.25cm}
\caption{Bode Plot of filter with an exponential chirp input of 1 amplitude from 0.1 to 100 Hz}
\label{Bode_Notch}
\vspace{-1cm}
\end{figure}

\subsubsection{Complex Multi-order Filter:}
The following transfer function:
\begin{equation*}
    H(s) = \frac{196.92s^3 + 21033.79s^2 + 427573.90s + 18317222.93}{s^3 + 382.16s^2 + 60851.34s + 3875784.59}
\end{equation*}
This transfer function can be digitally represented by the following input/output coefficient vectors with $f_l = 1000$ Hz.
\begin{align*}
    \hat{\textbf{a}} &= [1.7198E+02, -4.9816E+02,  4.8074E+02, -1.5455E+02] \\
    \hat{\textbf{b}} &= [2.6305E+00, -2.3162E+00,  6.8252E-01]
\end{align*}

The input/output and bode plot from this digital filter are as follows.
\begin{figure}[h] 
\includegraphics[width=0.9\linewidth]{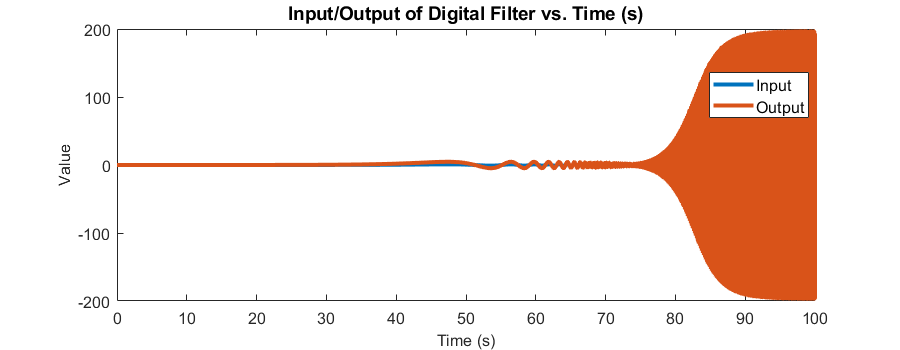}
\centering
\vspace{-0.25cm}
\caption{Input-Output plot of multi-order filter with an exponential chirp input of 1 amplitude from 0.1 to 100 Hz}
\label{IO_Multiorder}
\end{figure}

\begin{figure}[h] 
\includegraphics[width=0.9\linewidth]{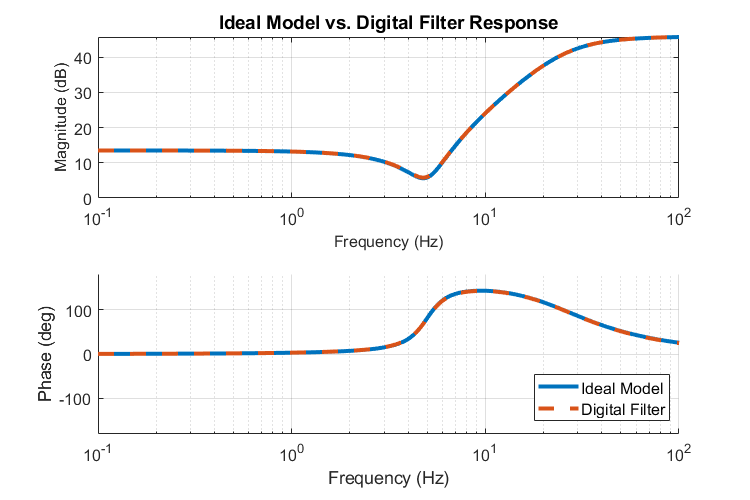}
\centering
\vspace{-0.25cm}
\caption{Bode Plot of filter with an exponential chirp input of 1 amplitude from 0.1 to 100 Hz}
\label{Bode_Multiorder}
\vspace{-1cm}
\end{figure}

\subsubsection{PID Controller:}
A PID Controller can be represented by the following causal (and unstable) transfer function:
\begin{align*}
    H(s) = (K_p + \frac{1}{s}K_i + \frac{\tau s}{s + \tau}K_d) = \frac{(K_p + K_d \tau)s^2 + (K_p\tau + K_i)s + K_i \tau}{s^2 + \tau s}
\end{align*}
where $K_p = 15.0$, $K_i = 2$, $K_d = 0.25$, and $\tau = 0.0035$ which simply acts as a low-pass filter on the derivative term for cleaner signals and causality. This transfer function can be digitally represented by the following input/output coefficient vectors with $f_l = 1000$ Hz.
\begin{align*}
    \hat{\textbf{a}} &= [1.5002E+01, -3.0002E+01,  1.5000E+01] \\
    \hat{\textbf{b}} &= [2.0000E+00, -1.0000E+00]
\end{align*}

The input/output and bode plot from this digital filter are as follows.
\begin{figure}[h] 
\includegraphics[width=0.9\linewidth]{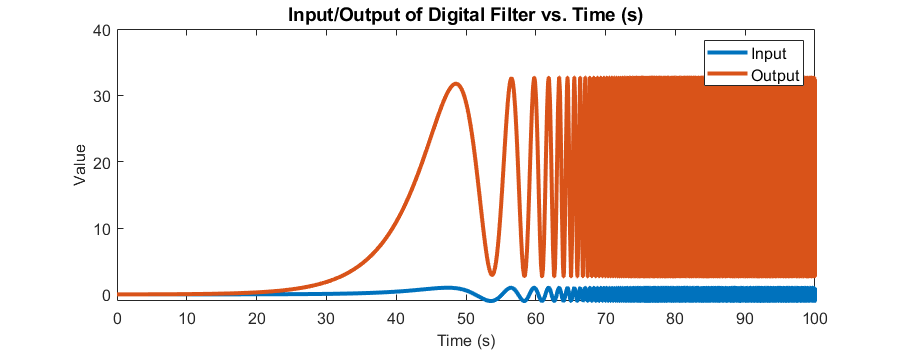}
\centering
\vspace{-0.25cm}
\caption{Input-Output plot of PID Controller with an exponential chirp input of 1 amplitude from 0.1 to 100 Hz}
\label{IO_PID}
\end{figure}

\begin{figure}[h] 
\includegraphics[width=0.9\linewidth]{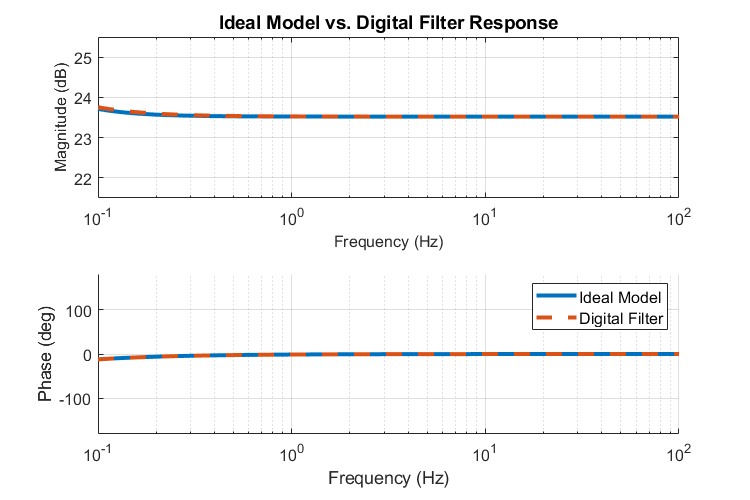}
\centering
\vspace{-0.25cm}
\caption{Bode Plot of filter with an exponential chirp input of 1 amplitude from 0.1 to 100 Hz}
\label{Bode_PID}
\vspace{-1cm}
\end{figure}

\subsubsection{Lead-Lag Compensator:} \label{sec:leadlag}
The following transfer function:
\begin{equation*}
    H(s) = K\frac{s+z}{s+p}
\end{equation*}
where $K = 10.0$, $z = 2\pi\cdot 1.0$, and $p = 2\pi\cdot 10.0$. This transfer function can be digitally represented by the following input/output coefficient vectors with $f_l = 1000$ Hz.
\begin{align*}
    \hat{\textbf{a}} &= [9.7259E+00, -9.6650E+00] \\
    \hat{\textbf{b}} &= [9.3908E-01]
\end{align*}

The input/output and bode plot from this digital filter are as follows.
\begin{figure}[h] 
\includegraphics[width=0.9\linewidth]{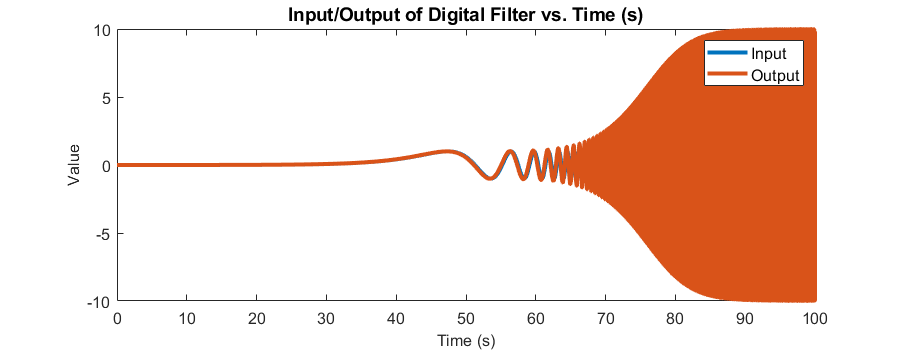}
\centering
\vspace{-0.25cm}
\caption{Input-Output plot of a Lead-Lag Compensator with an exponential chirp input of 1 amplitude from 0.1 to 100 Hz}
\label{IO_LeadLag}
\end{figure}

\begin{figure}[h] 
\includegraphics[width=0.9\linewidth]{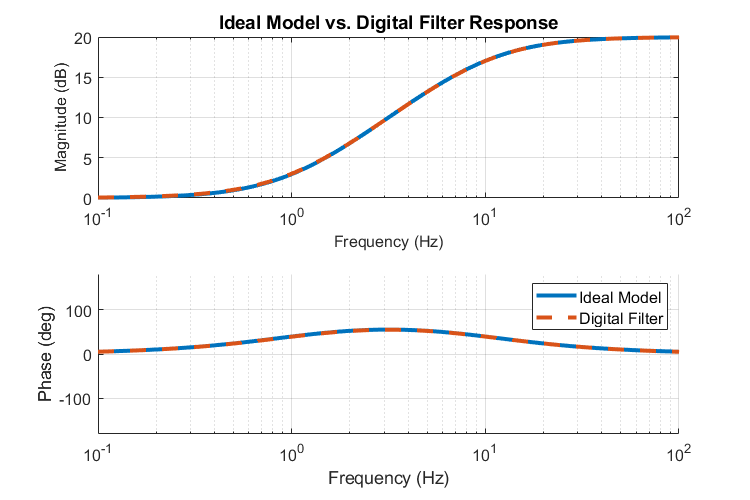}
\centering
\vspace{-0.25cm}
\caption{Bode Plot of filter with an exponential chirp input of 1 amplitude from 0.1 to 100 Hz}
\label{Bode_LeadLag}
\vspace{-1cm}
\end{figure}

\subsubsection{Heuristic for Avoiding High Outputs on Initialization}
As mentioned in Algorithm \ref{alg:cap}, a heuristic is added to set the $\mathbf{x}$ and $\mathbf{y}$ history vectors to the initial input. Without this heuristic, the first input and output history vectors can be initialized to zero and tend to have very high solutions at the start. This effect can be shown for the following example with a 2nd order Low-Pass Butterworth and Lead-Lag Controller, with equivalent transfer functions to Sections \ref{sec:order2lpbutter_filter} and \ref{sec:leadlag}, respectively. The input is a sinusoidal function of the form $x = \sin(2\pi\cdot100t) + 5$. 

\begin{figure}[h] 
\includegraphics[width=\linewidth]{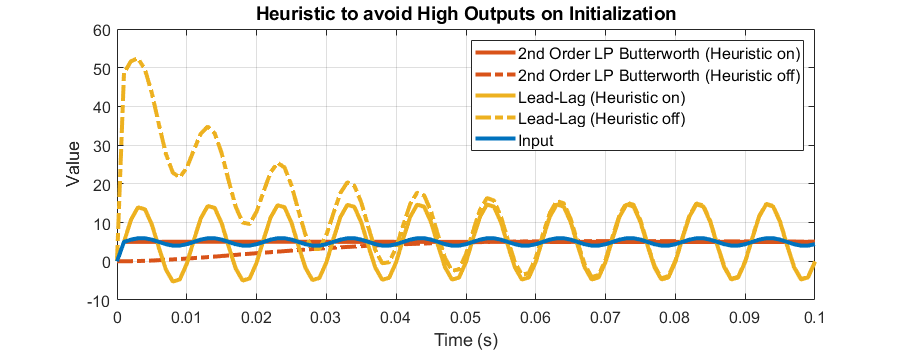}
\centering
\vspace{-0.25cm}
\caption{Start up heuristic to avoid large outputs for a 2nd order butterworth and lead-lag controller with an input of $x = \sin(2\pi\cdot100t) + 5$.}
\label{Bode_LeadLag}
\vspace{-1cm}
\end{figure}

\end{document}